\relax
\documentclass{llncs}
\usepackage{times}
\usepackage{helvet}
\usepackage{courier}
\usepackage{sectsty}
\sectionfont{\bfseries\Large\raggedright}

\usepackage[hyphens]{url}

\usepackage{xcolor}

\usepackage{amssymb,amsmath}

\usepackage{array,multirow}
\usepackage{tabularx, booktabs}
\newcolumntype{C}{>{\centering\arraybackslash}X}

\usepackage{graphicx,varwidth}
\usepackage[margin=6pt]{subfig}

\usepackage{wrapfig}

\begin{document}

\title{Knowledge Fusion via Embeddings from Text, Knowledge Graphs, and Images}

\author{Steffen Thoma \and Achim Rettinger \and Fabian Both}

\urldef{\mailsa}\path|{steffen.thoma,rettinger}@kit.edu|
\urldef{\mailsb}\path|fabian.both@student.kit.edu|

\institute{Karlsruhe Institute of Technology (KIT), Karlsruhe, Germany\\ \mailsa\\\mailsb}

\maketitle

\begin{abstract}
We present a baseline approach for cross-modal knowledge fusion. Different basic fusion methods are evaluated on existing embedding approaches to show the potential of joining knowledge about certain concepts across modalities in a fused concept representation. 
\end{abstract}

\section{Introduction}
\label{sec:intro}

This work intends to investigate the influence of modalities by means of tri-modal knowledge representations that fuse information from (1) text documents, (2) image collections and (3) large knowledge graphs (KGs). Intuitively, each one of these modalities provides complementing information. 

Based on this intuition we hypothesize that there is potential for a joint latent knowledge representation which is constructed from multiple embeddings. Such a representation might come closer to the human perception of concepts compared to representations extracted from a single modality alone.

To test this hypothesis, we propose an approach to integrate visual and textual latent representations with embeddings of KG concepts. When evaluating the resulting latent concept representations on standard similarity benchmarks, it indeed shows a higher correlation with the human notion of concept similarity than uni- or bi-modal representations.

This convincingly demonstrates the great potential of a joint latent knowledge representation constructed from multiple embeddings, as detailed in the following sections. First, we introduce existing uni-modal embeddings (Sec.~\ref{Sec:uni-modal}), before explaining how they are aligned (Sec.~\ref{sec:align}) and fused (Sec.~\ref{Sec:SharedSpace}). We demonstrate its potential on similarity benchmarks (Sec.~\ref{subsec:quantitativeAnalysis} and  Sec.~\ref{subsec:qualitativeAnalysis}) before we summarize our findings (Sec.~\ref{Sec:keyfindings}), discuss related work (Sec.~\ref{Sec:relatedwork}) and conclude (Sec.~\ref{Sec:conclusion}).

\section{Uni-Modal Vector Representations}
\label{Sec:uni-modal}

Latent vector representations of various types 
have become quite popular in recent years. The most common ones are latent \textit{textual representations}, which are also called \textit{word embeddings} or \textit{distributed word representations}. Created with unsupervised methods, they only rely on a huge text corpus as input. The information of co-occurrences with other words is encoded in a dense vector representation and by calculating the cosine similarity between two representations, a similarity score between two words is obtained. Examples for such word representations are \cite{bengio2006neural}, SENNA \cite{collobert2008unified}, hierarchical log-bilinear models \cite{mnih2008scalable}, word2vec \cite{mikolov2013efficient,mikolov2013distributed,mikolov2013linguistic}, and GloVe \cite{pennington2014glove}.

Similarly,  images can be encoded in a latent vector space. For \textit{image representations}, deep convolutional neural networks (CNN) have shown promising results in recent years. Deep CNNs transfer an image into a low dimensional vector space representation which for example can be used for image classification by applying a softmax function. The latent vector representation for images correspond to layers in the deep CNN before applying the softmax. For image classification with CNNs, Inception-V3 \cite{szegedy2015rethinking} which is used in TensorFlow \cite{abadi2016tensorflow} has shown good results on the ImageNet classification task \cite{ILSVRC15}.

The term 'Knowledge Graph' was coined by Google in 2012 and is since then used for any graph-based knowledge base, the most popular examples being DBpedia, Wikidata, and YAGO (see \cite{KG-survey} for a survey). 
Again, \textit{knowledge graph embeddings} can be learned on those graphs consisting of entities and typed predicates between entities and also abstract concepts. These entities and predicates can be encoded in a low dimensional vector space, facilitating the computation of probabilities for relations within the knowledge graph which can be used for link prediction tasks \cite{P6-rettinger2012mtsw-slfngkb}. Examples for learning latent vector representations of knowledge graphs are SE \cite{bordes2011learning}, RESCAL \cite {nickel2011three}, LFM \cite{jenatton2012latent}, TransE \cite{bordes2013translating}, SME \cite{bordes2014semantic}, HolE \cite{DBLP:conf/aaai/NickelRP16} and the SUNS framework \cite{huang2014scalable}.

\section{Tri-Modal Concatenated Concept Space}
\label{sec:align}

For obtaining a consolidated tri-modal shared space, a data representation that is aligned across multiple modalities is needed. Most existing bi-modal approaches rely on manually aligned documents. 
Since the manual creation of such datasets, containing a large number of text documents aligned with images and entities, is prohibitively expensive, we build on existing pre-trained uni-modal representations and align them across modalities after their respective training. This is achieved by relating modalities by more abstract concepts instead of the raw text, images and KG-instances.

Thus, we combine embeddings from three different modalities: \textit{textual}, \textit{visual} and \textit{knowledge graph}. We chose the well-established word2vec model for \textit{textual embeddings} and Inception-V3 for \textit{visual embeddings} since they have shown state-of-the-art results on their respective tasks and their pre-trained models are readily available online\footnote{Please note, that any other embedding, like the ones listed in the previous section, could be plugged-into our approach.}. For \textit{knowledge graph embeddings}, we could not find any suitable pre-trained model online so we trained representations ourselves. For that we chose the TransE model which has shown to be scalable to knowledge graphs with millions of vertices with good results \cite{krompass2015type}. For showing the better captured notion of human similarity within our approach, we evaluate on established human-made similarity datasets. Since the similarity datasets are based on word pairs, we align the representations to word-level:

\textbf{Word Matching of Word Embeddings} \quad 

The embeddings of word2vec do not need to be adjusted since they are already on word-level. Only the intersection of words representable by all modalities has to be identified.

\textbf{Concept Mapping of KG Embeddings} \quad 

The latent vectors of TransE are representing concepts in the DBpedia graph. Thus, one has to transform its representation to word-level. Each concept is uniquely addressable through a DBpedia URI and several labels (surface forms) are known which refer to this concept. For aligning the KG representations to word-level, we take for each concept the most commonly used surface forms for referring to this concept. 

\textbf{WordNet Mapping of Visual Features} \quad 

For visual representations, we use the images from \textit{ImageNet 1k} \cite{ILSVRC15}. Its 1000 ImageNet categories correspond to synsets from WordNet, where each of these synsets has a set of at least 1300 images for the respective topic. By combining all image representations for a given synset, one can obtain a visual representation for the synset. Alike to \cite{kiela2014learning} we combine these image representations by taking the max-value for each vector index as we have also noticed that these yield better results compared to the mean values. Additionally, more abstract synset representations can be built from synset representations: Utilizing the WordNet hierarchy, for example an embedding of \textit{`instrument'} can be created by combining embeddings of \textit{`violin'} and \textit{'harp'}. Therefore, we build hierarchical subtrees in WordNet for each synset not contained in \textit{ImageNet 1k}. All synset representations in such a subtree, with a visual representation from \textit{ImageNet 1k}, are then combined to form the abstract synset representation. 
In total, we abstract 396 additional synset representations. The alignment of the synset representations to the word-level can be performed with WordNet lexemes (words) since lexemes are assigned to at least one synset in WordNet. In the end, we extract 2574 lexeme representations by utilizing our 1396 synset representations. The intersection with word2vec and TransE leads to an aligned tri-modal concept space with 1523 concepts. Representations from all modalities can be concatenated for each shared concept so that fusion techniques for the resulting concept space can be applied.

\section{Shared Cross-modal Concept Space}
\label{Sec:SharedSpace}

For obtaining the shared space from the vector representations, the modalities have to be fused. Apart from simple concatenation we build on a set of existing methods like SVD and PCA for dimensionality reduction. In addition, we propose a normalization and weighting scheme for embeddings from multiple modalities. Our tri-modal concept space of $n$ different word-phrases is represented in three matrices: text $T$, knowledge graph $G$, and visual $V$. For combination techniques, we use the whole information of all three modalities and define matrix $M \in \mathbb{R}^{(t+g+v) \times n}$ as the vertically stacked matrices of $T$, $G$, and $V$.
The three latent vector representations vary drastically in their dimensions: \textit{Visual representations} tend to have more than 1000 dimensions while \textit{knowledge graph representations} typically have around 50 to 100 dimensions. Thus, the representations with higher dimensionalities tend to dominate the combination techniques. 
To circumvent this dimensionality bias, \cite{goikoetxea2016single} combine only vectors of equal size. Furthermore, the value range of features can differ, depending on the underlying training objective and method. To address these problems we tested pre-processing steps, comprising \textit{normalization (N)} of each column vector of $T$, $G$, and $V$ to unit length as well as \textit{weighting (W)} of the normalized matrices $T$, $G$, and $V$ individually with weights $w_T$, $w_G$, and $w_V$ before stacking. Thus, we take into account that certain representations might contain more useful information or might be better trained.

\textbf{AVG} \quad The averaging method uses the cosine similarity of all three modalities which are calculated separately. By averaging these three values, we get a combined similarity measure which is also robust with respect to different vector dimensionalities.

\textbf{CONC} \quad The vector representations can be combined by concatenation of single representations and the similarity is calculated by cosine similarity between two concatenated vectors. The similarities of the following techniques are also calculated with cosine similarity.

\textbf{SVD} \quad Singular value decomposition factorizes the input matrix $M$ into three matrices such that $M = U \Sigma V^T$. $U$ and $V$ are unitary matrices and $\Sigma$ is a diagonal matrix with the singular values of $M$ in descending order on its diagonal. By taking the first $k$ columns of $U$ and the $k$ biggest singular values of $\Sigma$, we get a new combined $k$-dimensional representation: $M \leftarrow M_k = U_k \Sigma_k$.

\textbf{PCA} \quad Principal Component Analysis uses an orthogonal transformation to convert the correlated variables into linearly uncorrelated variables. Fixing the number of uncorrelated principal components results in a projection into a lower dimensional vector space. By taking the principal components with the highest variance, we create a representation with the most distinctive features. We also tested canonical correlation analysis (CCA) but in our tests PCA always performed superior which is consistent with \cite{goikoetxea2016single}. Thus, we omit further tests with CCA. 

\section{Empirical Evaluation}
\label{Sec:eval}

In our quantitative and qualitative assessments, we use the pre-trained representations for text and images and newly trained knowledge graph representations. Textual representations were obtained from word2vec\footnote{\url{https://code.google.com/archive/p/word2vec/}}, its vectors have 300 dimensions and were trained with negative sampling on the Google News text corpus containing about 100 billion words. 
For visual representations, the Inception-V3 model\footnote{\url{http://download.tensorflow.org/models/image/imagenet/inception-v3-2016-03-01.tar.gz}}, pre-trained on the \textit{ImageNet 1k} classification task, was applied to compute representations with 2048 dimensions.
Knowledge graph representations were obtained with the TransE model \cite{bordes2013translating} which we created ourselves by running TransE on the DBpedia knowledge graph. We trained TransE with local closed word assumption for type constraints, rank=50, gamma=0.3, learningrate-embeddings=0.2 and learningrate-parameters=0.5 on the DBpedia version 3.9. We made all used embeddings 
available online\footnote{\label{goo.gl}\url{https://people.aifb.kit.edu/sto/TriM1523}}.

\subsection{Quantitative Similarity Assessment}
\label{subsec:quantitativeAnalysis}

For evaluation, we utilize various word similarity datasets. The limiting factor are verbs, abstract words and named entities (e.g. persons) for which there is no visual representation available. Thus, we provide the subsets of \textit{MEN} \cite{bruni2014multimodal}, \textit{WS-353} \cite{finkelstein2001placing}, \textit{SimLex-999} \cite{hill2015simlex}, and \textit{MTurk-771} \cite{halawi2012large} which are covered in the combined space online. As aggregated measure, we also report the average performance over all evaluation datasets, weighted by their respective size. Table~\ref{completeVSsubset} confirms that the subsets of the word evaluation datasets are similarly difficult to solve compared to the original datasets. 

\begin{table}[ht]
\centering
\caption{Spearman rank correlation of subsets and complete datasets for word2vec.}
\label{completeVSsubset}
\begin{tabularx}{0.7\linewidth}{l|CCCCC|}
     & MEN & WS- & SimLex- & MTurk- & weighted \\ 
     &  & 353 & 999 & 771 & $\varnothing$ \\ \hline 
complete  & 0.762 & 0.700 & 0.442 & 0.671 & 0.682 \\ 
subset  & 0.740 & 0.707 & 0.423 & 0.594 & 0.669 \\ \hline
\end{tabularx}
\end{table}

In Table~\ref{overview}, Spearman rank correlation on all subsets for raw stacking, normalization (N) and weighting (W) are reported. Combination methods with dimensionality reduction use 100 dimensions. For raw stacking, none of the combination methods is significantly better than the single textual representation on the \textit{MEN}, \textit{MTurk-771} and \textit{WS-353} subset. 

\begin{table}[ht]
\centering
\caption{Spearman rank correlation on subsets of evaluation datasets.}
\label{overview}
\begin{tabularx}{0.7\linewidth}{l|CCCCC|}
     & MEN & WS- & SimLex- & MTurk- & weighted \\
     & & 353 & 999 & 771 & $\varnothing$ \\ \hline 
Visual  & 0.619 & 0.526 & 0.522 & 0.308 & 0.546 \\ 
Textual  & 0.740 & 0.707 & 0.423 & 0.594 & 0.669 \\ 
KG  & 0.423 & 0.425 & 0.246 & 0.275 & 0.375 \\ \hline
AVG  & 0.763 & 0.691 & 0.436 & 0.524 & 0.672 \\ 
CONC & 0.620 & 0.521 & 0.522 & 0.310 & 0.547 \\ 
SVD  & 0.740 & 0.646 & 0.588 & 0.325 & 0.643 \\ 
PCA  & 0.483 & 0.392 & 0.591 & 0.272 & 0.455 \\ \hline 
AVG-N  & 0.763 & 0.691 & 0.436 & 0.524 & 0.672 \\ 
CONC-N & 0.763 & 0.691 & 0.436 & 0.524 & 0.672 \\ 
SVD-N  & 0.747 & 0.606 & 0.397 & 0.500 & 0.644 \\ 
PCA-N  & 0.691 & 0.427 & 0.336 & 0.415 & 0.571 \\   \hline 
AVG-W  & 0.807 & 0.715 & 0.585 & 0.594 & 0.733 \\
CONC-W  & 0.806 & 0.726 & 0.586 & 0.589 & 0.732 \\ 
SVD-W  & \textbf{0.847} & 0.687 & \textbf{0.616} & \textbf{0.618} & \textbf{0.762} \\
PCA-W  & 0.836 & \textbf{0.760} & 0.586 & 0.568 & 0.751 \\  \hline
\end{tabularx}
\end{table}

Normalized representations allow for a fixed combination ratio, resembling an equal weight of information from all modalities. 
We also conducted additional experiments with different dimension parameters for SVD and PCA. Thereby, we concluded 
that 100 dimensions are sufficient to encode the useful information for the word similarity task. However, combination methods with normalization (N) are also not significantly and consistently outperforming the single textual representation as shown in Table~\ref{overview}.

Representations from different modalities are most likely not equally informative for the word similarity task and hold different complementary information. We use weighting (W) of representations after normalization in order to control the proportion of information induced by each representation. With grid search and a step size of 0.05 we investigated the modality composition on the weighted average of all evaluation sets. 
The optimal weights for ($w_T$, $w_G$, $w_V$) are: AVG (0.15, 0.10, 0.75), CONC (0.25, 0.15, 0.6), SVD (0.25, 0.1, 0.65), and PCA (0.3, 0.05, 0.65). Although some of the weighting schemes only include small proportions of the KG representations, extracted complementary information from KGs improves performance significantly. Fig.~\ref{fig_weighting} illustrates that the combination of the three fused and weighted modalities produces better results than the single representations. Weighted combination methods substantially outperform uni- and bi-modal embeddings while best results are obtained with PCA and SVD. For combination methods without dimensionality reduction, a similar weighting optimum and performance can be observed for CONC and AVG.

\subsection{Qualitative Similarity Assessment}
\label{subsec:qualitativeAnalysis}

\begin{figure*}[!t]
\centering
\subfloat[AVG]{\includegraphics[width=0.398\linewidth]{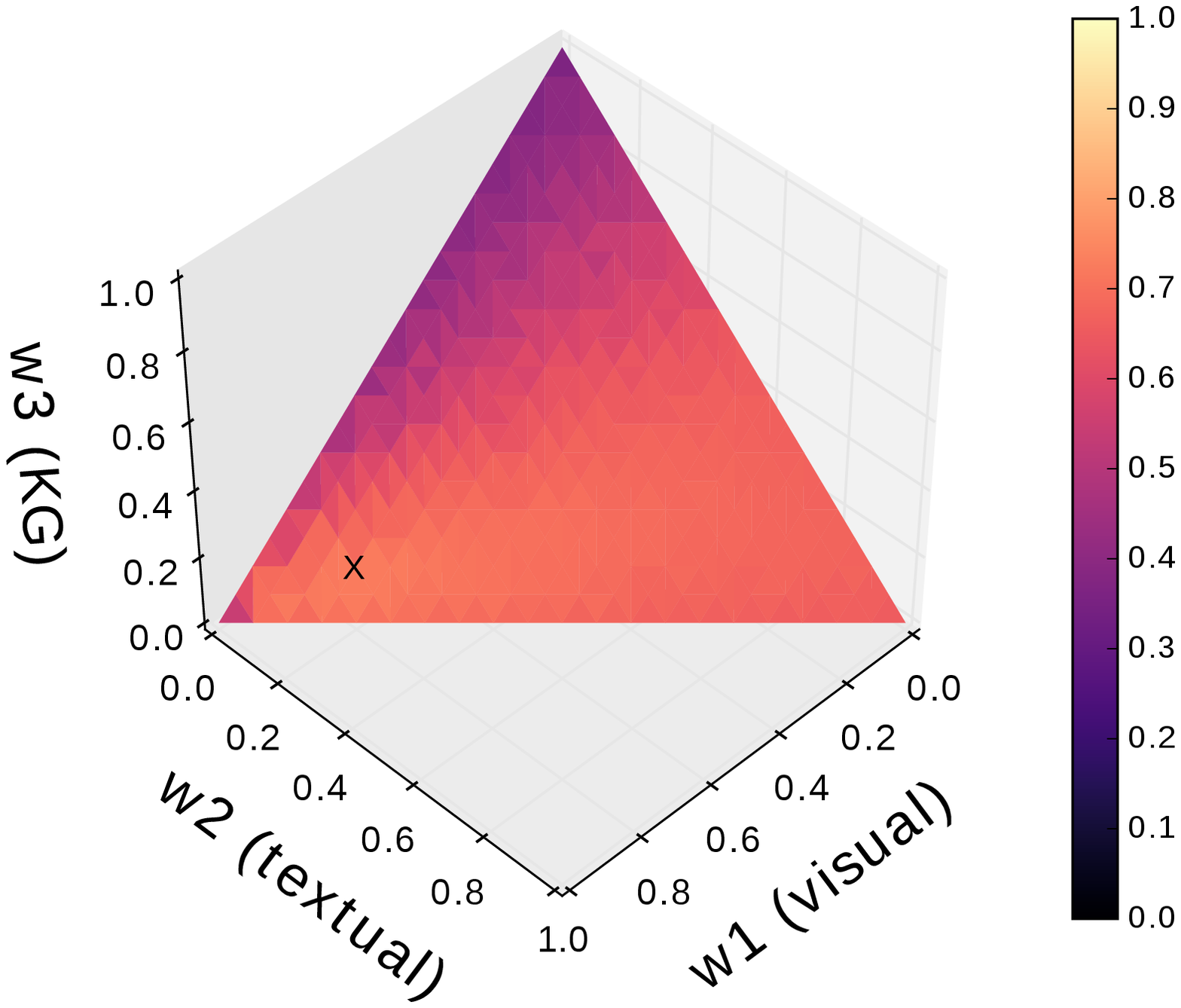}%
\label{w_AVG}}
\hfil
\subfloat[CONC]{\includegraphics[width=0.398\linewidth]{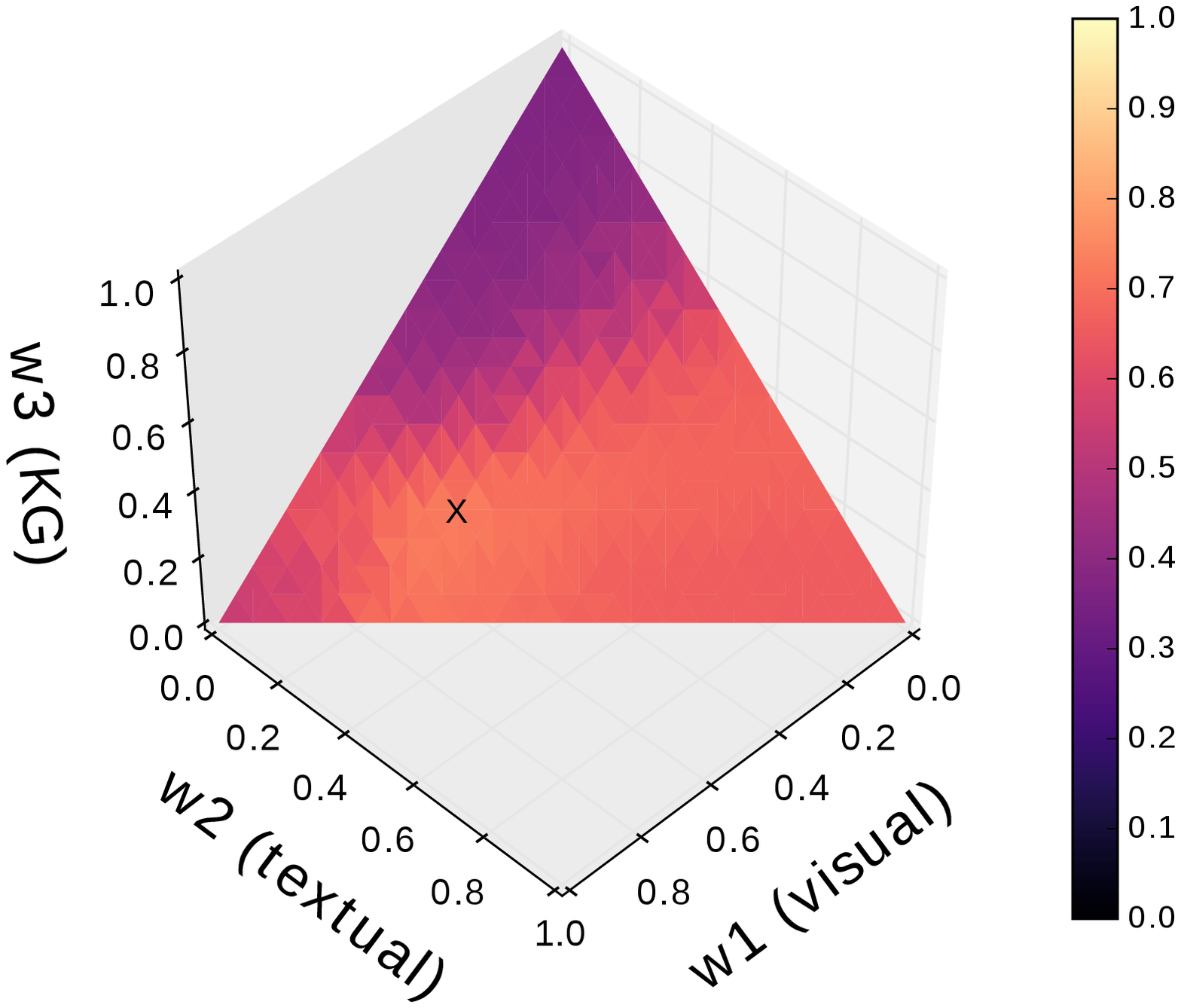}%
\label{w_CAT}}
\hfil
\newline
\subfloat[PCA 100 dim]{\includegraphics[width=0.398\linewidth]{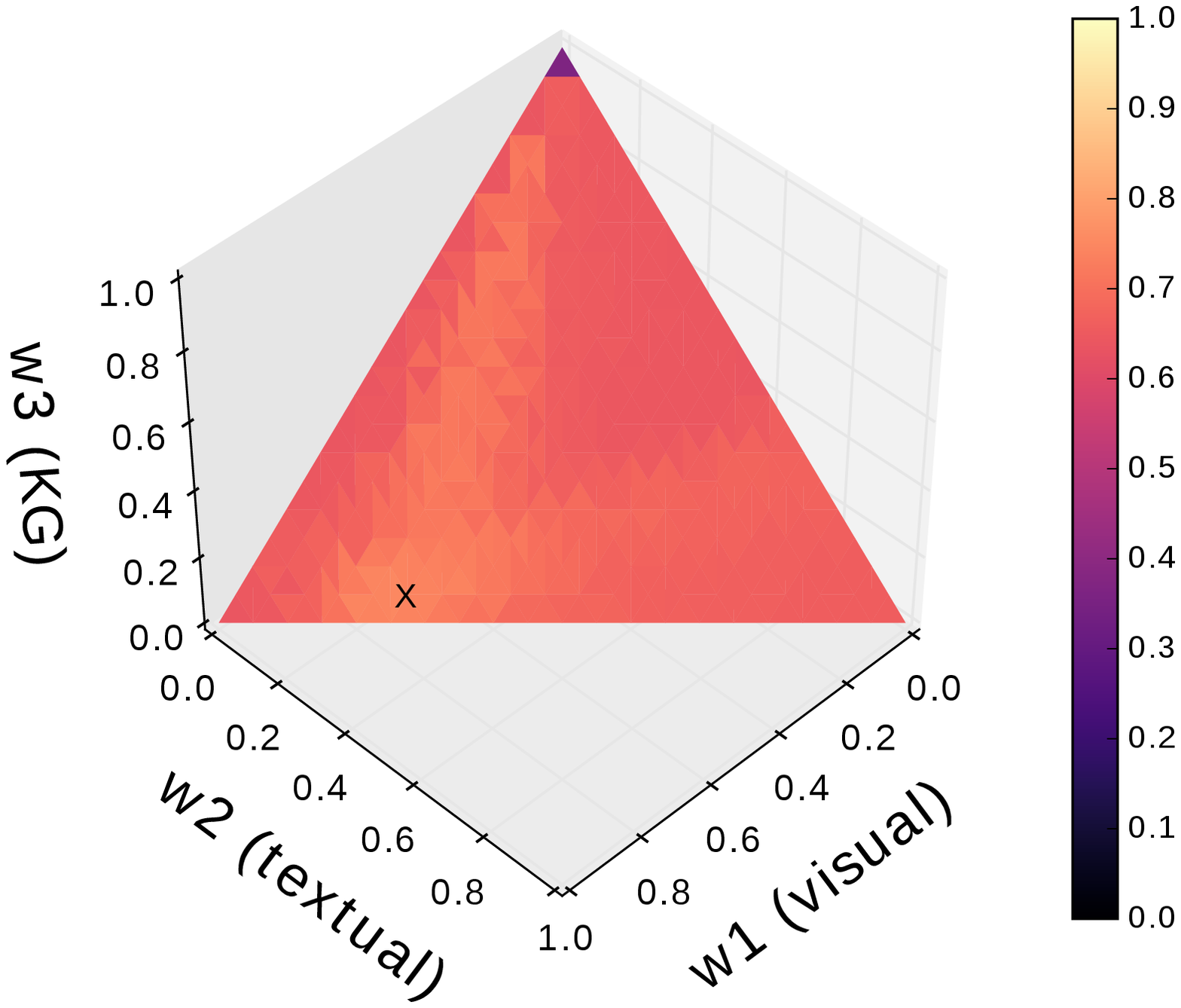}%
\label{w_PCA}}
\hfil
\subfloat[SVD 100 dim]{\includegraphics[width=0.398\linewidth]{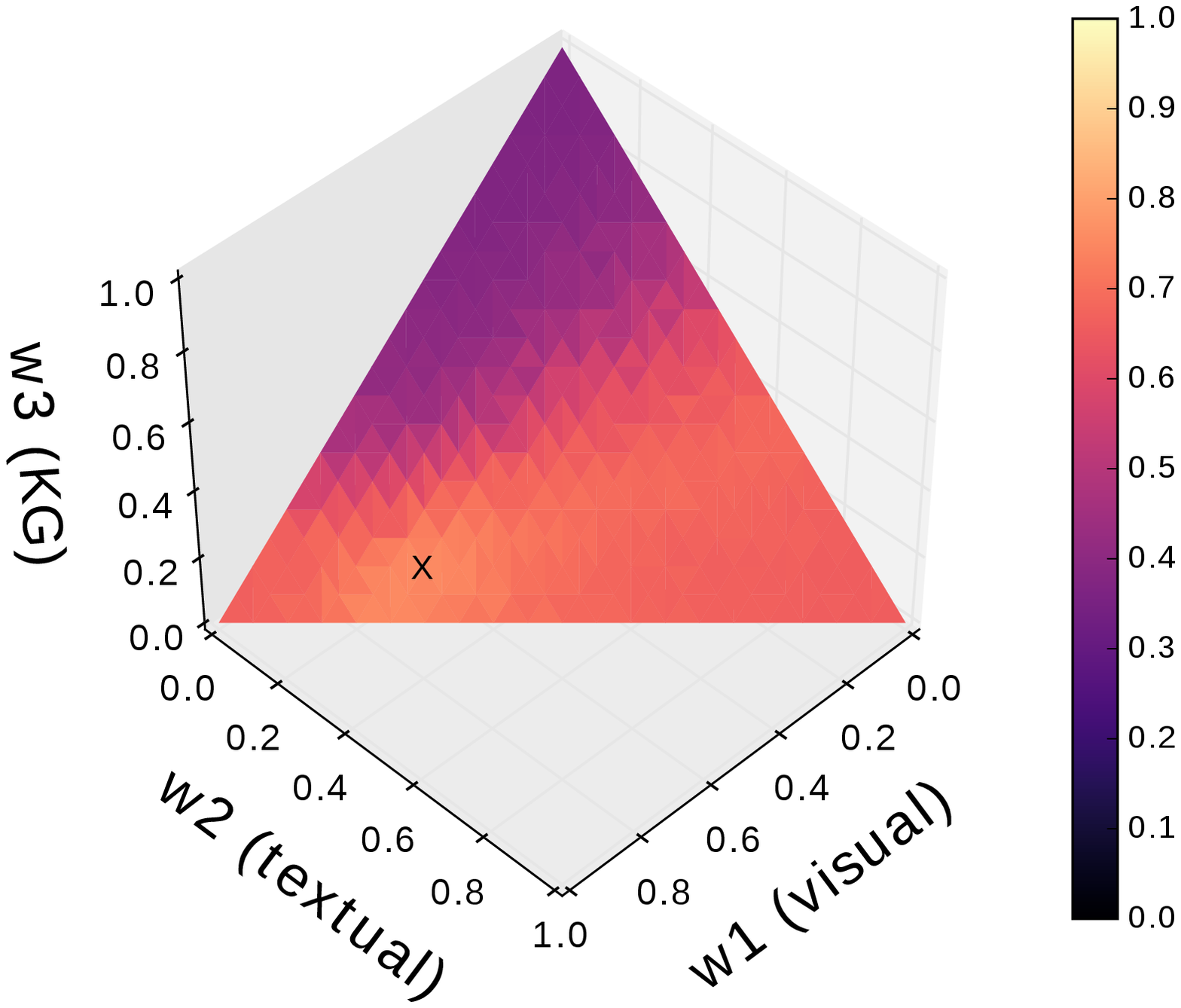}%
\label{w_SVD}}
\caption{Weighting with normalization. Shown are average plots over all three evaluation datasets, the colorbar indicates Spearman rank correlation and the black cross marks the optimum.}
\label{fig_weighting}
\end{figure*}

We conducted experiments for \textit{MEN}, \textit{WS-353}, \textit{SimLex-999}, and \textit{MTurk-771} with normalization: In Fig.~\ref{fig_weighting} you can see the weighted average of Spearman rank correlation scores for different weightings between normalized textual, KG, and visual representations. For combination via AVG and CONC shown in Fig.~\ref{w_AVG} and \ref{w_CAT}, we observed similar behavior on all evaluation datasets in terms of weighting optimum and structure.

Further, we analyze PCA and SVD for 100 dimensions. It turns out that weighting after normalization is even more crucial for those methods as it is for the AVG and CONC method. Especially PCA exploits information from KG representations with very low weight. Nonetheless, the combined representation of all three modalities is significantly better than only a combination of bi-modal representations.

Optimal weights always include all three modalities. Further experiments with different TransE model parameterizations revealed that this finding is not depending on a specifically trained TransE embedding, but attributes to information extracted from the knowledge graph. Thus, we can improve concept representations from other modalities with complementary information encoded in TransE, Inception-V3, and word2vec embeddings. 

\subsection{Key Findings}
\label{Sec:keyfindings}

\begin{itemize}

\item We successfully exploited pre-trained concept representations from different domains. This enables the re-usability of years of research in Computer Vision, Natural Language Processing and Semantic Web. Plugging new models into our approach and incorporating further modalities is straightforward.

\item In our empirical evaluation, we could verify that there exists complementary information in \textit{visual}, \textit{textual} and \textit{KG embeddings}, which can be exploited by fusing all representations.


\item The performance of embeddings in the word similarity task were improved significantly. The shared representations always performed best if information from all three modalities was included. 


\end{itemize}

\section{Related Work on Fusion of Learned Representations}
\label{Sec:relatedwork}

Recently, several researchers have tried to transfer learned knowledge from one task to another or to combine different approaches. In image classification, it is important that also new images can be classified such that visual representations from one image classification task can be transferred to another with different classes. To this end, \cite{oquab2014learning} learn and transfer mid-level image representations of CNNs. \cite{kiela2014learning} test the combination of visual and textual representations via vector stacking 
which is similar to \cite{silberer2014learning} who use a stacked auto-encoder to combine visual and textual input. In contrast to our approach they only evaluated simple vector stacking and neither evaluated more sophisticated combination techniques nor the incorporation of structured resources like KGs. 

In contrast, \cite{goikoetxea2016single} uses textual information from a text corpus and WordNet. For this purpose WordNet is transferred to text by performing random walks on the synset hierarchy and hereby storing the traversal path to text \cite{goikoetxea2015random}. But, they neither use visual representations nor do they work with the information of an expressive KG directly. The transformation of a traversal path to text might lose characteristics of the underlying graph structure which is why we used latent vector representations from an explicit KG model, learned on a complete KG.

Our approach also goes beyond current retrofitting ideas like \cite{faruqui2015retrofitting}. They adjust learned word embeddings by incorporating information from lexical databases. Firstly, we do not slightly adapt one representation but learn a new combined representation. Secondly, we use much more information from a large expressive KG (DBpedia) instead of a smaller lexical database. Lastly, we also use visual information.

The closest work to our word-level alignment in concept space is \cite{rothe2015autoextend}\footnote{Please note, that they did not consider any combinations with visual or KG embeddings.}. They used autoencoders with rank 4 weight tensors to create vector representations for synsets and lexemes in WordNet for which there was no learned vector representation before. They achieve this by treating a word and a synset as the sum of its lexemes.

\section{Conclusion and Future Work}
\label{Sec:conclusion}

This work demonstrates the potential of mining meaningful concept representations from multi-modal data sources. Our approach builds on the aligned embeddings extracted from 1.3 million images, a text corpus of 100 billion words and 7 million KG concepts by fusing all aligned embeddings in one \emph{shared cross-modal concept space}. 
We optimized for creating the most holistic representation of a concept by fusing knowledge from all available modalities. This is limited to a small set of concepts since there are only few concepts which are available in all modalities so far. Currently visual embeddings pose a bottleneck compared to the exhaustive concept spaces of textual and KG embeddings.

We introduced a two-step process of word-level alignment and fusion for different modalities. We demonstrated that the fused embeddings of all three modalities can outperform uni- and bi-modal embeddings. This confirms our hypothesis, that each modality contributes complementing information and thus, a fusion of modalities in one shared concept space can yield a more holistic view. This also advocates the importance of semantically meaningful embedding spaces. 
Further, our results indicate that a shared multi-modal representation comes closer to the human notion of the concept which we demonstrated quantitatively with similarity benchmarks. 
It appears that knowledge representations come closer to a human-like perception of concepts if all modalities are considered. 

Our findings open up a large number of future research directions by raising fundamental questions. First and foremost, it became obvious that knowledge representations in general and knowledge graphs in particular can benefit greatly from integrating latent semantics from multiple modalities. Our approach is a first attempt to achieve that. However, a pressing issue that remains for future work is how this can be scaled to a larger number of concepts.
Beyond that, there are numerous cross-disciplinary research challenges in areas like sensory neuroscience, philosophy of perception or multimodality research that could benefit from advances in the area of learned cross-modal knowledge representations and vice versa.


\bibliographystyle{splncs03}
\bibliography{bibliography}

\end{document}